\documentclass[letterpaper, 10 pt, conference]{ieeeconf}  

\IEEEoverridecommandlockouts                              

\usepackage[backend=biber,
            hyperref=true,
            url=false,
            isbn=false,
            doi=false,
            backref=false,
            style=ieee,
            natbib=true,
            citestyle=numeric-comp,
            sorting=nyt,
            mincitenames=1,
            maxcitenames=1,
            block=none]{biblatex}

\renewcommand{\bibfont}{\small}
\usepackage{graphics}
\usepackage[pdftex]{graphicx}
\usepackage{wrapfig}
\DeclareGraphicsExtensions{.pdf,.png,.jpg}
\usepackage{epsfig}
\usepackage[font={small}]{caption}
\usepackage{subcaption}
\usepackage[rightcaption]{sidecap}
\usepackage{pbox}

\usepackage{bigstrut}
\usepackage{makecell}

\usepackage{mathtools}
\usepackage{amssymb,amsmath}
\usepackage{gensymb} 
\usepackage{nicefrac}       
\numberwithin{equation}{section} 
\usepackage{algorithm}
\usepackage{algpseudocode}

\usepackage{array} 
\usepackage{tabularx}
\usepackage{multirow}
\usepackage{multicol}
\usepackage{booktabs}
\usepackage{tabulary}

\usepackage[utf8]{inputenc}
\usepackage{units}
\usepackage{bm}
\usepackage{xspace}
\usepackage{flushend}
\usepackage{balance} 
\usepackage{csquotes}
\usepackage{makeidx}
\usepackage{blindtext}
\usepackage[dvipsnames]{xcolor}

\usepackage{autolabtools}

\let\labelindent\relax
\usepackage{enumitem}

\usepackage{tikz}
\usetikzlibrary{backgrounds}

\usepackage{ragged2e}
\usepackage{soul} 
\usepackage{subfiles} 

\usepackage{url}
\makeatletter
\g@addto@macro{\UrlBreaks}{\UrlOrds}
\makeatother

\usepackage{hyperref}
\hypersetup{
    colorlinks=true,
    linkcolor=black,
    citecolor=black,
    filecolor=cyan,
    urlcolor=black
}

\renewcommand{\baselinestretch}{1.0}



\newcommand{\algname}{HOUSTON}

\title{\LARGE \bf
Learning to Localize, Grasp, and Hand Over \\Unmodified Surgical Needles}

\author{Albert Wilcox$^{*}$, Justin Kerr$^{*}$, Brijen Thananjeyan, Jeffrey Ichnowski, \\ Minho Hwang, Samuel Paradis, Danyal Fer, Ken Goldberg
  \\ \scriptsize{* equal contribution} \\
\thanks{The AUTOLab at UC Berkeley (\href{mailto:automation@berkeley.edu}{automation@berkeley.edu})}%
\thanks{\version{3.1}}
}

\begin{document}

\maketitle

\begin{abstract}
Robotic Surgical Assistants (RSAs) are commonly used to perform minimally invasive surgeries by expert surgeons. However, long procedures filled with tedious and repetitive tasks such as suturing can lead to surgeon fatigue, motivating the automation of suturing. As visual tracking of a thin reflective needle is extremely challenging, prior work has modified the needle with nonreflective contrasting paint. As a step towards automation of a suturing subtask without modifying the needle, we propose HOUSTON: Handoff of Unmodified, Surgical, Tool-Obstructed Needles, a problem and algorithm that uses a learned active sensing policy with a stereo camera to localize and align the needle into a visible and accessible pose for the other arm. To compensate for robot positioning and needle perception errors, the algorithm then executes a high-precision grasping motion that uses multiple cameras. In physical experiments using the da Vinci Research Kit (dVRK), \algname{} successfully passes unmodified surgical needles with a success rate of $96.7\%$ and is able to perform handover sequentially between the arms $32.4$ times on average before failure. On needles unseen in training, \algname{} achieves a success rate of $75-92.9\%$. To our knowledge, this work is the first to study handover of unmodified surgical needles. See
\url{https://tinyurl.com/houston-surgery}
for additional materials.
\end{abstract}

\section{Introduction}
\label{sec:introduction}



Robotic Surgical Assistants (RSAs) currently rely on human supervision for the entirety of surgical tasks, which can consist of many very repetitive subtasks such as suturing. Automation of surgical subtasks may reduce surgeon fatigue~\cite{yip2017robot}, with initial results in surgical cutting~\cite{thananjeyan2017multilateral,murali2015learning}, debridement~\cite{seita_icra_2018,murali2015learning}, suturing~\cite{sen2016automating,superhuman_dict,thananjeyan2019safety,chiu2020bimanual,saeidi_suturing_icra_2019,extraction_needles_2019,automated_needle_pickup_2018,improved_knots_case_2013}, hemostasis~\cite{ritcher_bloodflow_2020}, and peg transfer~\cite{paradis2020intermittent,hwang2020applying,hwang2020efficiently,auto_peg_transfer_2015}.

This paper considers automation of the bimanual regrasping subtask~\cite{chiu2020bimanual} of surgical suturing, which involves passing a surgical needle from one end effector to another. This handover motion is performed in between stitches during suturing, and is a critical step, as accurately positioning the needle in the end effector affects the stability of its trajectory when guided through tissue. Because varying cable tension in the cables driving the arms causes inaccuracies in motions, a high precision task such as passing a needle between the end effectors is challenging~\cite{mahler2014case,hwang2020applying,paradis2020intermittent,hwang2020efficiently,peng2020real}.

The task is also difficult because 3D pose information is critical for successfully manipulating needles, and surgical needles are challenging to perceive with RGB or active depth sensors due to their reflective surface, thin profile~\cite{kollar2021simnet}, and self-occlusions (Figure~\ref{fig:tiers}). Prior work has mitigated this by painting needles~\cite{sen2016automating,chiu2020bimanual} and using color segmentation, but this solution is not practical for clinical use.

\begin{figure}[t]
    \centering
    \includegraphics[width=\columnwidth]{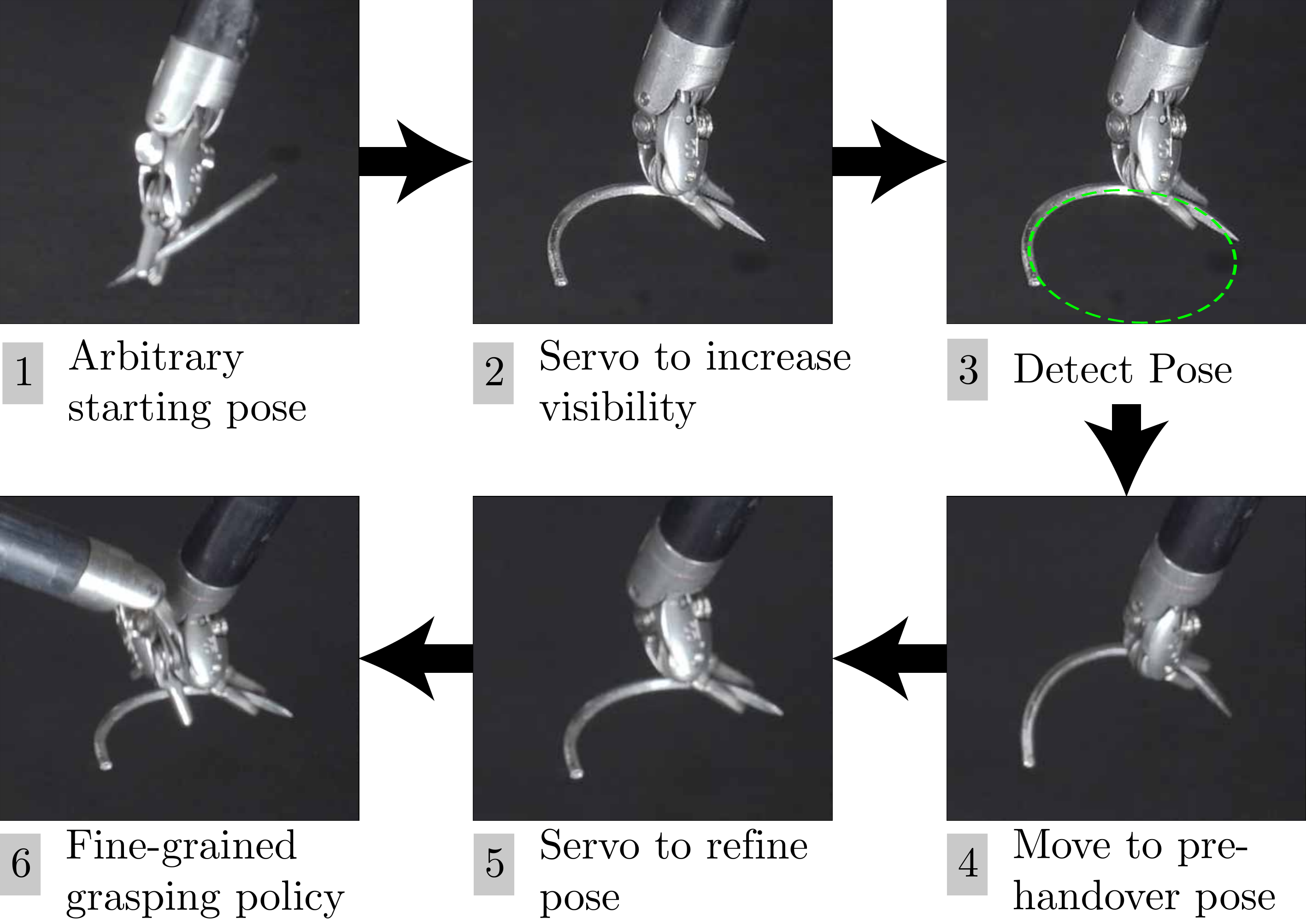}
    \caption{\textbf{Algorithm overview}: The algorithm first servos the needle into a position that is easily visible to the stereo camera to produce a high confidence pose estimate. Using this estimate, it coarsely reorients the needle towards the grasping arm, and then refines this iteratively to correct for positioning errors. Finally, it executes a learned visual servoing grasping policy to complete the handover.}
    \label{fig:algo}
\end{figure}

To manipulate unmodified surgical needles, we combine recent advances in deep learning, active sensing, and visual servoing. We present HOUSTON: Handoff of Unmodified, Surgical, Tool-Obstructed Needles, a problem and algorithm for using stereo vision with coarse and fine-grained control policies (Figure~\ref{fig:algo}) to sequentially localize, orient, and handover unmodified surgical needles. We present a localization method using stereo RGB with a deep segmentation network to output a point cloud of the needle in the workspace. This point cloud is used to define a coarse robot policy that uses visual servoing to reorient the needle for handover in a pose that is visible to the cameras and accessible by the other end effector. However, due to inaccuracies of robot positioning and the perception system, further corrections may be necessary. We train a fine robot policy from a small set of human demonstrations to perform these subtle but critical corrections from images for unmodified surgical needles.

This paper makes the following contributions:
\begin{enumerate}
    \item A perception pipeline using stereo RGB to accurately estimate the pose of surgical steel needles in 3D space, enabling needle manipulation without active depth sensors and painted needles.
    \item A visual servoing algorithm to perform coarse reorientation of a surgical needle for grasping.
    \item A needle grasping policy that performs fine control of the needle learned from a small set of human demonstrations to compensate for robot positioning and needle sensing inaccuracies.
    \item Combination of the pose estimator (1), the servoing algorithm (2) and needle controller (3) to perform bimanual surgical needle regrasping, where physical experiments on the da Vinci Research Kit (dVRK)~\cite{dvrk2014} suggest a success rate of $96.7\%$ on needles used in training, and $75-92.9\%$ on needles unseen in training. On sequential handovers, \algname{} successfully executes 32.4 handovers on average before failure.
\end{enumerate}

\section{Related Work}
\label{sec:related_works}


\subsection{Automation in Surgical Robotics}
Automation of surgical subtasks is an active area of research with a rich history. Prior literature has studied automation of tasks related to surgical cutting~\cite{thananjeyan2017multilateral,murali2015learning,krishnan2019swirl}, debridement~\cite{murali2015learning,kehoe2014autonomous}, hemostasis~\cite{ritcher_bloodflow_2020}, peg transfer~\cite{hwang2020applying,hwang2020efficiently,paradis2020intermittent}, and suturing~\cite{sen2016automating,thananjeyan2019safety,chiu2020bimanual,extraction_needles_2019,automated_needle_pickup_2018,saeidi_suturing_icra_2019}. While automated suturing has been studied in prior work~\cite{saeidi_suturing_icra_2019,sen2016automating}, suturing without modifications such as painted fiducial markers is an open research problem. Recent work studies robust and general approaches to specific subproblems within suturing, including the precise manipulation of surgical needles during suturing from needle extraction~\cite{extraction_needles_2019} to bimanual regrasping~\cite{chiu2020bimanual}, which is the focus of this work.

Needle manipulation is also studied by~\cite{extraction_needles_2019}, where the approach studies the extraction of needles from tissue phantoms to compute robust grasps of the needle even in self-occluded configurations. Bimanual needle regrasping was studied in detail by~\citet{chiu2020bimanual}, with impressive results on simulation-trained policies that take needle end effector poses as input. We extend their problem definition to consider multiple handoffs of unmodified needles and end effectors, which requires perception of the needle and robot pose from images without color segmentation. Needle grasping has also been studied using visual servoing policies in~\citet{automated_needle_pickup_2018}, where the needle is painted with green markers to track its position during closed loop visual servoing.
\citet{varier2020collaborative} study tabular RL policies for needle regrasping in a discretized space in a fixed setup with known needle pose and experiments without the needle on the dVRK. The experiments in~\cite{varier2020collaborative} suggest that value iteration-trained policies can mimic expert trajectories used for inverse reinforcement learning. In contrast, we present an algorithm compatible with significantly varying initial needle and gripper poses using only image observations. We additionally present many physical experiments with a needle, evaluating the success rate and speed of the algorithm.




\subsection{Visual Servoing, and Active Perception}
Visual servoing (VS) is a popular technique in robotics~\cite{hutchinson1996tutorial,kragic2002survey}, and has recently been applied to compensate for surgical robot imprecision in the surgical peg transfer task~\cite{paradis2020intermittent}. While classical VS approaches typically make use of hand-tuned visual features and known system dynamics~\cite{chaumette2006visual,caron2013photometric}, recent work proposes learning end-to-end visual servoing policies from examples~\cite{levine2018learning,QT-Opt}. To reduce the need for tuned features and dynamics models and also reduce the number of training samples required to create a robust VS policy for bimanual needle regrasping, we present a hybrid approach that combines coarse motion planning with fine control, where a learned VS policy is only used in parts of the task where high precision is required. This framework, called intermittent visual servoing (IVS), was studied in detail by \cite{paradis2020intermittent}, where the system switches between a classical trajectory optimizer and imitation learning VS policy based on the precision required at the time. Inspired by this technique, we present an IVS approach to bimanual needle regrasping, that combines coarse perception and planning with fine VS control. Because this task requires reasoning about depth across several directions, we present a multi-view VS policy that learns to precisely hand over the needle based on several camera views.

Active perception is a popular technique with many variations to localize objects prior to manipulation by maximizing information gain about their poses~\cite{mihaylova2002comparison,salaris2017online,whitehead1990active,bajcsy1988active,arruda2016active}. This has been studied in the context of robot-assisted surgery, where the endoscope position is automatically adjusted via a policy learned from demonstrations to center the camera focus on inclusions during surgeon teleoperation. In this work, we actively servo the needle to highly visible poses to maximize the accuracy of its pose estimate. This is most similar to~\cite{arruda2016active}, where the authors propose an algorithm to actively select views of a grasping workspace to uncover enough information about unknown objects to plan grasps.


\section{Problem Formulation}
\label{sec:prob_form}

The \algname{} problem extends and generalizes the previous problem definition from~\citet{chiu2020bimanual} to include unmodified needles, occlusion, and multiple handoffs.



\subsection{Overview}
\begin{figure}[t]
    \centering
    \vspace{2mm}
    \includegraphics[width=\columnwidth]{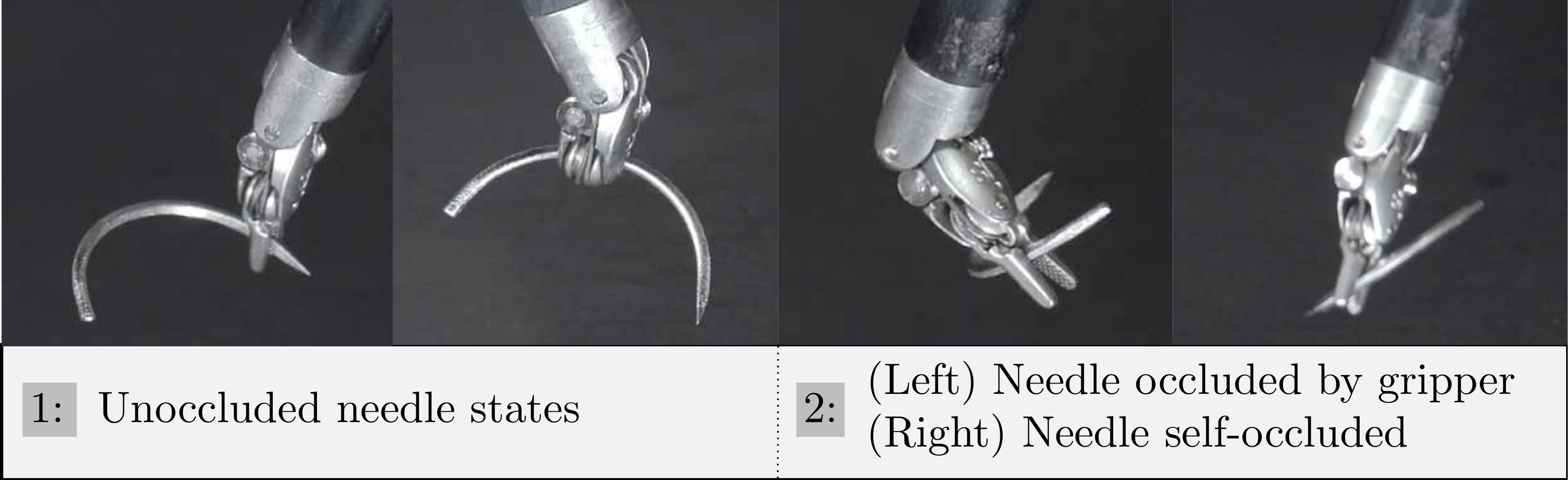}
    \caption{\textbf{Visible and occluded configurations: }The left two frames depict needle orientations that are easily identifiable. However, the needle frequently reaches states that are occluded by the gripper or self-occluded, which makes estimating its state challenging.}
    \label{fig:tiers}
\end{figure}

In the \algname{} problem, the surgical robot starts with a curved surgical needle with known curvature and radius grasped by one gripper and must accurately pass it to the other gripper and back.
This is challenging due to the needle's reflective surface, thin profile, and pathological configurations~\cite{extraction_needles_2019,chiu2020bimanual} as depicted in Figure \ref{fig:tiers}. Once the needle is successfully passed to the other end effector, it is passed back to the first end effector. This process is repeated $N_\mathrm{max}$ times, or until the needle is dropped. We also consider a special case of this problem, the single-handover version, in which $N_\mathrm{max}=1$.

\subsection{Notation}
Let $p_L(t)$ and $p_R(t)$ denote the poses of the left and right grippers, respectively, at discrete timestep $t$ with respect to a world coordinate frame. The needle has pose $p_N(t)$ with respect to the world frame. Observations of the workspace are available via RGB images from a stereo camera, $I_{L}(t)$ and $I_{R}(t)$, or overhead monocular RGB images $I_O(t)$ from an RGB camera. The left and right cameras in the stereo pair have world poses $p_{\rm st, left}$ and $p_{\rm st, right}$, respectively. The overhead camera has pose in world frame $p_{\rm O}$. Each trial starts with the needle in the left gripper and ends when the needle is dropped. Additionally, the trial terminates if no successful handoff occurs in $\tau_{\rm max}$ timesteps.

At timestep $t$, the algorithm is provided observation $y(t)$ which contains images from the sensors: $y(t) = \left(I_L(t), I_R(t), I_O(t)\right)$. The algorithm outputs a target pose and jaw state for each end effector $u(t) = ((p_{L}(t+1), \varphi_L(t+1)), (p_{R}(t+1), \varphi_{R}(t+1)))$, where $\varphi_L(t+1) \in \{0, 1\}$ indicates the whether the left jaw is closed at timestep $t+1$.


\subsection{Assumptions}\label{subsec:assumptions}
\begin{figure}[t]
    \centering
    \vspace{2mm}
    \includegraphics[width=\columnwidth]{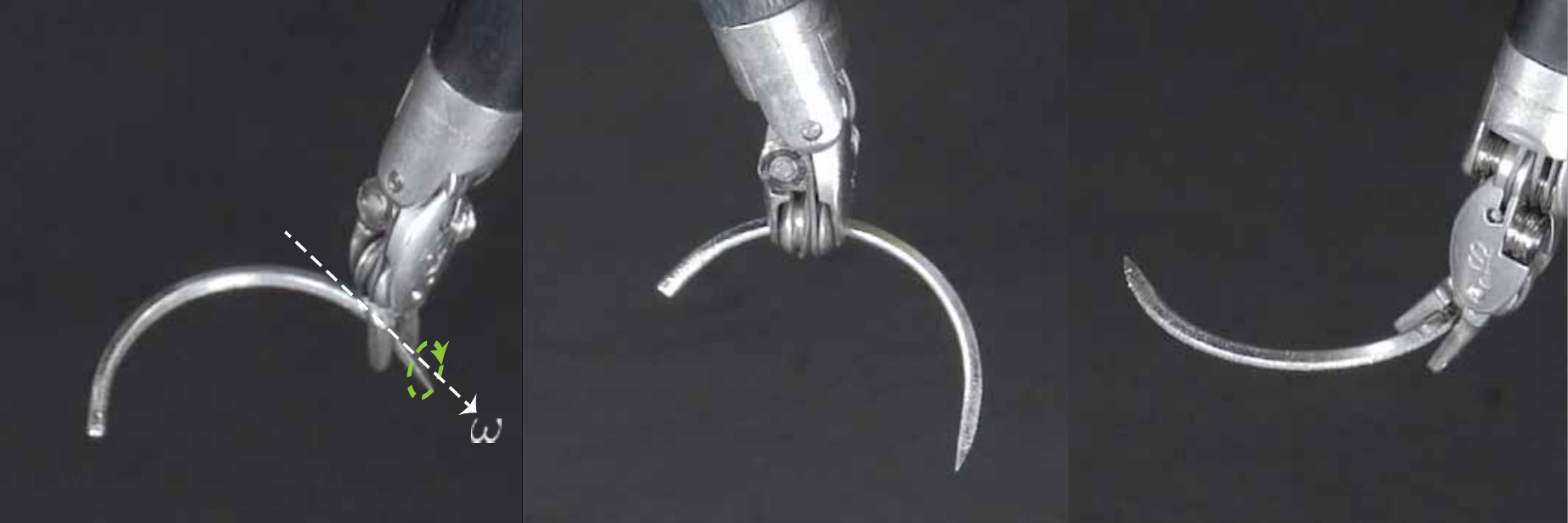}
    \caption{\textbf{Needle Reset Degrees of Freedom:} We vary the starting configurations of the needle relative to the gripper in three degrees of freedom as described in \ref{subsec:assumptions}. \textbf{Left:} We rotate the needle into 7 discretized states about the axis $\omega$ tangent to the needle. \textbf{Middle:} An example holding the middle 30\degree inward from the tip. \textbf{Right:} An example where the needle's arc faces towards the camera.}
    \label{fig:my_label}
\end{figure}

In order to deterministically evaluate \algname{} policies in a wide variety of needle configurations, we discretize the needle-in-gripper pose possibilities by choosing a number of categories across three degrees of freedom:
\begin{enumerate}
    \item The needle's curve can face either towards or away from the camera, providing 2 possibilities
    \item The gripper may hold the needle either at the tip, or 30\degree~inwards following the curvature of the needle. This degree of freedom has 2 possibilities.
    \item The rotation of the needle about the tangent line $\omega$ to the point the gripper intersects ranges from $0\degree$ to $180\degree$, and is discretized into 7 bins as in Figure~\ref{fig:my_label}.
\end{enumerate}
This gives a total of 28 possible needle configurations and we perform a grid search over these possibilities. We chose these configurations to be representative of those seen in suturing tasks post-needle extraction.

While the robot encoders provide an estimate of the gripper poses $p_L(0)$ and $p_R(0)$, the precise needle pose is unknown due to cabling effects of the arms. We assume access to a stereo RGB pair in the workspace, an overhead RGB camera, and the transforms between the coordinate frames of these cameras and the robot arms.



\subsection{Evaluation Metrics}
We evaluate \algname{} by recording: \textbf{\emph{i})} the number of successful handoffs in a multi-handoff trial, \textbf{\emph{ii})} the success rate per arm of single handoffs beginning from each configuration in \ref{subsec:assumptions}, and \textbf{\emph{iii})} the average time for each handoff.

\section{\algname{} Algorithm}
\label{sec:method}


\algname{} uses active stereo visual servoing with both a coarse-motion and fine-motion learned policy for the bimanual regrasping task.



\subsection{Phase 1: Active Needle Presentation}\label{subsec:reorientation}

\begin{figure}
     \centering
     \vspace{2mm}
     \includegraphics[width=\columnwidth]{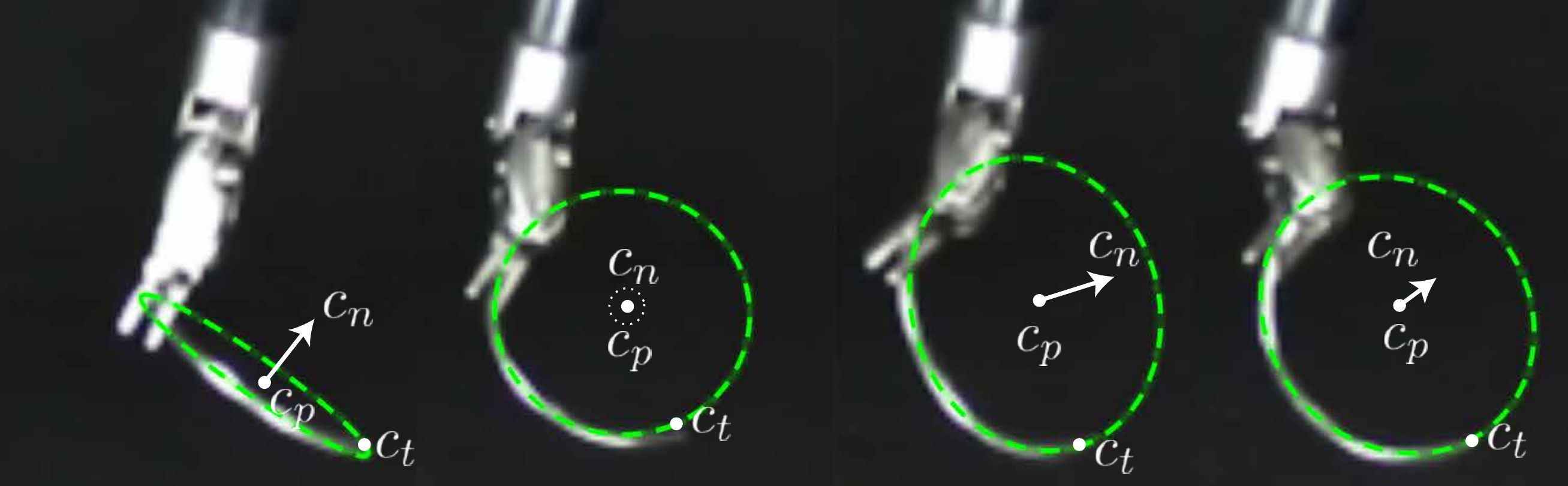}
     \caption{\textbf{Acquisition Stage Rollout:} Rollout of the phase described in \ref{subsec:reorientation}. Images are taken from the left camera and cropped to the gripper, with the circle observation projected in green. The pose goal is one in which the needle faces towards the camera. Note how uncertainty in the second image is resolved in later images as the needle reaches a more observable configuration.}
        \label{fig:vs_reorient_rollout}
\end{figure}

In the first phase, the algorithm repositions the needle to a pose where the other arm can easily grasp it without collisions and the cameras can clearly view it. Throughout execution of the coarse policy, we parameterize the needle as a circle of known radius, and measure its state in world frame as a center point $c_p$, normal vector $c_n$, and a needle tip point $c_t$ as shown in Figure~\ref{fig:vs_reorient_rollout}. Active Needle Presentation consists of two stages: \textit{needle acquisition} and \textit{handover positioning}. The needle acquisition stage moves the needle to maximize visibility, and the positioning stage uses visual servoing to move the needle into a graspable state.
\subsubsection{Needle State Estimation}

\begin{figure}[t]
     \centering
     \vspace{2mm}
     \includegraphics[width=\columnwidth]{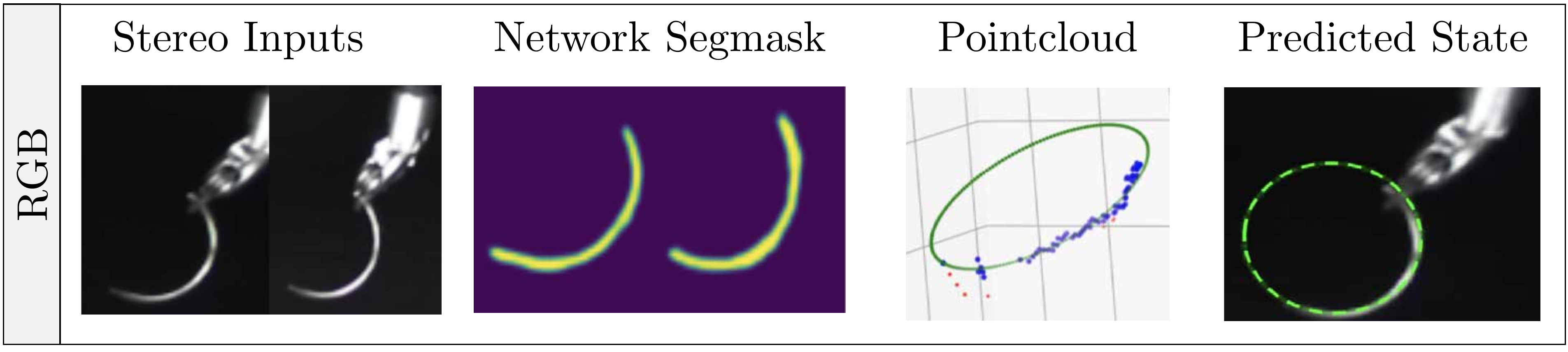}
     \caption{\textbf{Needle state estimation:} Execution of the stereo RGB pipeline on a highly visible needle. The network takes in raw stereo images as input, producing segmasks of the needle. The triangulated segmasks produce a point-cloud to which a circle is fit with RANSAC (3rd panel). Inliers are shown in blue, the best-fit circle in green, and outliers in red. The resulting observation reprojected into the left image is shown in the final panel.}
        \label{fig:depth_needle_detection}
\end{figure}

The state estimator passes stereo images into a fully convolutional neural network that is trained to output segmentation masks for the needle in each image. See the project website for architecture details. It computes a distance transform of the segmentation mask to label each pixel with its distance to the nearest unactivated pixel. Next, it finds peaks in the distance transform along horizontal lines in each image, which correspond to points near the center of activated patches. It then triangulates all pairs of peaks along each horizontal line in the images to obtain a point-cloud as in Figure \ref{fig:depth_needle_detection}. Because this may triangulate outlier points from the gripper or incorrectly match points on different parts of the needle, RANSAC is applied to filter out incorrect point correspondences and returns the final predicted needle state. At each iteration, it samples a set of 3 points, to which a plane is fit. Each subset of 3 points generates 2 candidate circles in the plane, corresponding to the two circles which pass through one of the pairs. RANSAC uses an inlier radius of 1mm and runs for 300 iterations.

The network is first trained on a dataset of 2000 simulated stereo images of randomly-placed, textured and scaled floating needles and random objects~\cite{calli2017yale} above a surface plane generated in Blender 2.92.
Lighting intensity, size, and position, and stereo camera position are also randomized. The segmentation network is fine-tuned on a dataset of 200 manually-labeled images of the surgical needle in the end effector. Training a network on a PC with an NVIDIA V100 GPU takes 3 hours, and fine tuning takes 10 minutes.

\subsubsection{Visual servoing}
\algname{} uses Algorithm~\ref{alg:reorientation} to compute updates for visual servoing in both the needle presentation and the handover positioning phases. Algorithm~\ref{alg:reorientation} is a fixed point iteration method that uses a state estimator to iteratively visually servo to a target state. Similar to first and second order optimization algorithms, it computes a global update based on its current state and iterates until the computed update is zero. In each iteration, it queries the current 3D state estimate of the needle and then computes an update step in the direction of the target state. To compensate for estimation errors due to challenging needle poses, this process is repeated at each iteration until the algorithm converges to a local optimum within a pose error tolerance.

During \textbf{needle acquisition}, the arm moves to a home pose, then rotates around the world $z$ and $x$ axes, stopping when the state estimator observes at threshold number of inlier points $n_{\rm thresh}=20$ from RANSAC circle fitting. During trials, at most 2 consecutive rotations sufficed to resolve the state to this degree. After this initial acquisition, we apply Algorithm~\ref{alg:reorientation} to align $c_n$ towards the left stereo camera position $p_{\rm st, left}^{pos}$, with $d$ defined as a rotation about the axis $c_n\times(p^{pos}_{\rm st, left}-c_p)$. Once clearly presented to the camera, we measure $c_t$ by choosing the inlier point from the circle fitting step which is furthest from the gripper in 3D space.

Subsequently, during \textbf{handover positioning}, we compute inverse kinematics to move $c_p$ towards the center of the workspace with $c_t$ pointing towards the other gripper and $c_n$ orthogonal to the table plane. This flat configuration is critical for the grasping step, since the dVRK arm is primarily designed for top-down grasps near the center of its workspace. Because the arm only has 5 rotational degrees of freedom, we use a numerical IK solver from~\cite{tracik} and attempt to find a configuration minimizing rotational error within a $6\times6\times8$ cm tolerance region on end effector translation. After moving to this pose, we repeat Algorithm~\ref{alg:reorientation}, with $d$ defined as a rotation aligning $c_t$ towards the other gripper and $c_n$ orthogonal to the table. We calculate IK to a configuration with needle curvature towards the camera and one with curvature away from the camera, then pick the configuration which minimizes rotational error to the goal.






\begin{figure}
    \centering
    \vspace{2mm}
    \includegraphics[width=\columnwidth]{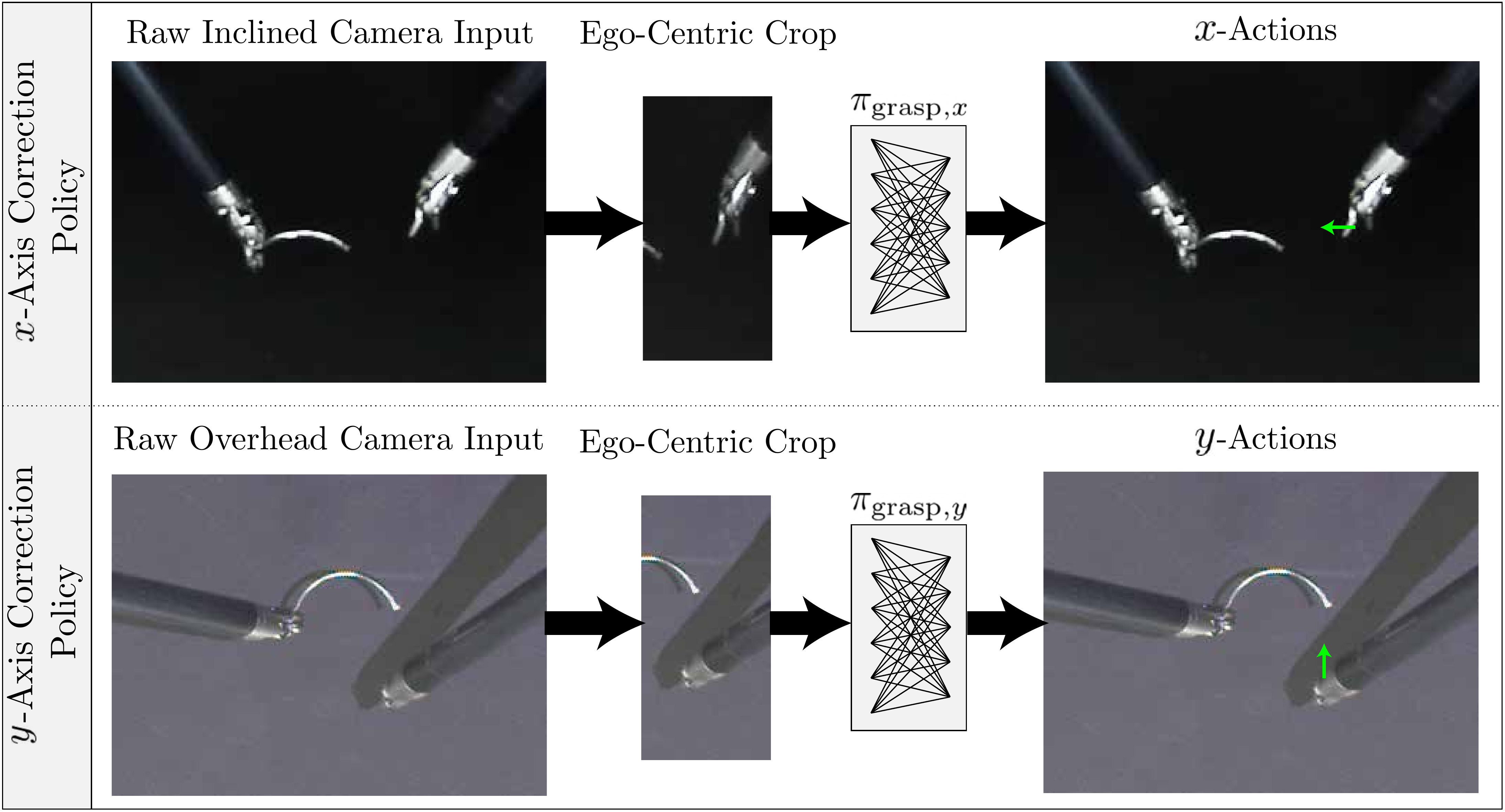}
    \caption{\textbf{Fine-grained grasping policy:} We split corrective actions along the $x$- and $y$-axes and learn two corresponding policies $\pi_{\mathrm{grasp},x}$ and $\pi_{\mathrm{grasp},y}$. Each policy begins by performing an ego-centric crop by projecting the gripper's kinematically calculated approximate location into the input image and cropping around it. Then, it feeds the cropped image through a neural network to predict a correction direction.}
    \label{fig:handover}
\end{figure}

\subsection{Phase 2: Executing a Grasping Policy}\label{subsec:ivs_grasping}
After the active presentation phase described in Section~\ref{subsec:reorientation} and the pose of the needle is relatively accurately known and accessible to the grasping arm, we execute a grasping policy $\pi_{\rm grasp}$ to grasp the needle. However, the needle pose estimate after the first phase may not be perfect, so we must visually servo to compensate for these errors when grasping. Even if the needle pose was perfectly known, reliable grasping of a small needle is still challenging due to the positioning errors of the robot, which are a result of its cable-driven arms~\cite{paradis2020intermittent,hwang2020efficiently,hwang2020applying,seita_icra_2018,peng2020real,mahler2014case}. The policy splits corrective actions between the $x$- and $y$-axes with two sub-policies, $\pi_{\mathrm{grasp},x}$ and $\pi_{\mathrm{grasp},y}$. Each policy uses RGB inputs ego-centrically cropped around the grasping arm, with the $x$- and $y$-axis policies using $140 \times 200$ pixel crops from the inclined camera and $70 \times 200$ pixel crops from the overhead camera respectively. The cropping forces the policy to condition based on the relative position of the gripper and needle without the ability to overfit to texture cues from other parts of the scene. The fine-grained grasping policy and image crops are displayed in Figure~\ref{fig:handover}. We ablate different design choices for the grasping correction policy and also present open-loop grasping results in Section~\ref{sec:experiments_and_results}.

The grasping subpolicy $\pi_{\mathrm{grasp},y}$ is a neural network classifier that outputs whether the grasping arm should move in the $+y$ (down in the crop) or $-y$ (up in the crop) direction. $\pi_{\mathrm{grasp},x}$ is trained similarly to output whether the grasping arm should move in the $+x$ or $-x$ directions. The policies are trained by collecting offline human demonstrations through two methods: 1) we sample poses for arms in the workspace such that needle orientation is perturbed by $10\degree$ about each axis, then move the robot to a good grasping position via a keyboard teleoperation interface. 2) we execute the pre-handover positioning routine and position the robot in the desired grasp location by hand, after which the robot autonomously iterates through offsets in the $\pm x$ and $\pm y$ directions, labeling actions according to the offset from goal position. We experimentally find that separating the policy across two axes significantly improves grasp accuracy (Section~\ref{sec:experiments_and_results}). A separate grasping policy is trained for each arm on 100 demonstrations each. Each demonstration takes 5-10 actions and each dataset takes about an hour to collect. The policies $\pi_{\mathrm{grasp},x}$ and $\pi_{\mathrm{grasp},y}$ are each represented by voting ensembles of 5 classifiers, each of which have three convolutional layers and two fully connected layers. Details about model architectures are located in the project website.

During policy execution, we iteratively sample actions first from $\pi_{\mathrm{grasp},x}$, then $\pi_{\mathrm{grasp},y}$, waiting for each to converge before continuing. We multiply the action magnitude by a scalar $\beta_{decay}=0.5$ every time the network outputs the opposite action of the previous timestep. Servoing terminates when action magnitude decays to under $0.2$mm. This enables implicit convergence to the goal without explicitly training the policy to stop. After $x$ and $y$ convergence, we execute a simple downward motion of 1cm to grasp the needle.

\begin{algorithm}[htb!]
  \caption{Presentation Visual Servoing Policy}
  \label{alg:reorientation}
\begin{algorithmic}[1]
\Require State estimator $\Phi$, target needle state $x^{\rm targ}_N$, arm grasping needle $a \in \{L, R\}$, current time $t$, number of iterations $N_{\rm max}$, tolerance $\epsilon$, distance metric $d$.
\For{$i = 0; i< N_{\rm max}; i = i + 1$}
    \State Predict needle state $\hat{x}_N(t+i) = \Phi(o(t+i))$
    \State Compute gripper update $p_{\delta} = d(\hat{x}_N(t+i), x^{\rm targ}_N)$
    \State Update arm $a$ pose: $p_{a}(t+i+1) = p_{a}(t+i) + p_\delta$
    \If{$|p_\delta|<\epsilon$}
        \State \textbf{break}
    \EndIf
\EndFor

\end{algorithmic}
\end{algorithm}

\section{Physical Experiments} \label{sec:experiments_and_results} 



The experiments aim to answer the following question: how efficient and reliable is \algname{} compared to baseline approaches? We also perform several ablation studies of method components in this section.


\subsection{Baselines}\label{subsec:comparisons}
To evaluate the method for the task of active needle presentation, we compare to the following baselines:
\begin{itemize}
    \item \textit{Depth-based presentation:} Instead of using the stereo RGB network to detect needle pose, we use a depth image-based detection algorithm to detect the needle pose and servo it to the flat grasping pose. This method takes the depth image from the built in depth calculation from the stereo camera, $I_{D}(t)$, as input, masks the gripper out of the depth image using the dVRK's forward kinematics, then performs a volume crop around the end effector and fits a circle of known radius to the points using RANSAC to extract the state.
    \item \textit{No-Sim-Data:} This is an ablation of the RGB stereo segmentation network that is only trained on the small dataset of real data.
\end{itemize}

\begin{figure}
    \centering
    \vspace{2mm}
    \includegraphics[width=\columnwidth]{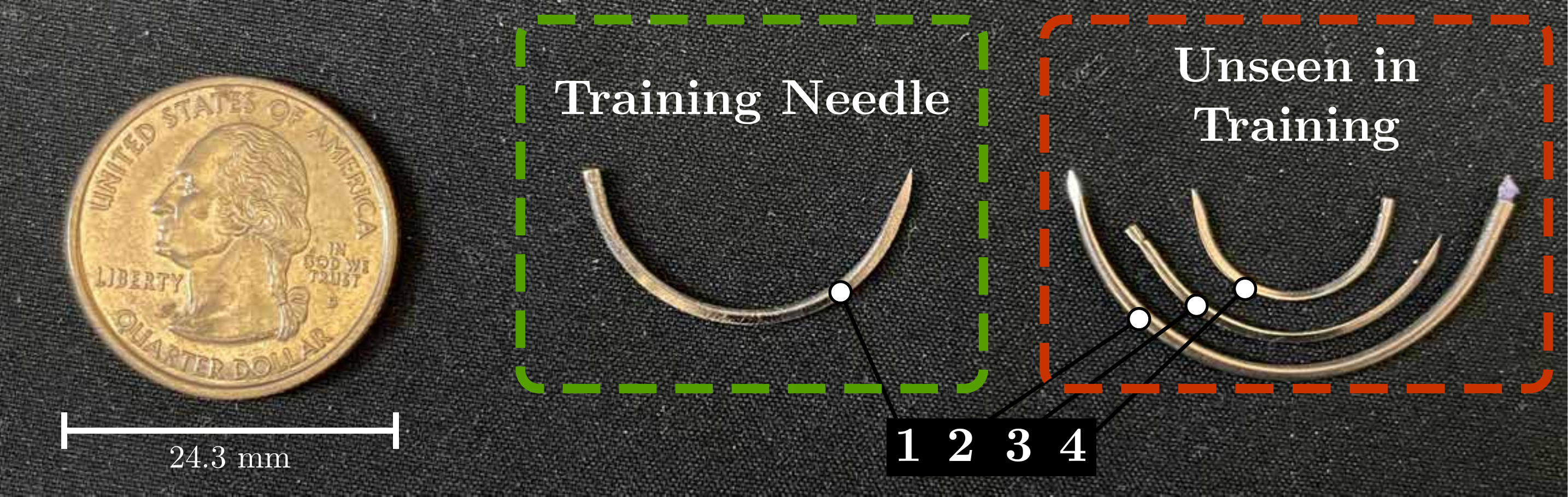}
    \caption{\textbf{Needles used:} The needles are shown with a coin for scale. All models were trained using needle 1, while needles 2, 3 and 4 were also used for testing. Needles 1 and 3 have a radius of 1.25cm, and needles 2 and 4 have radii of 1.75cm and 0.75cm}
    \label{fig:needle_sizes}
\end{figure}

\begin{table*}[t]
\vspace{0.1in}
\caption {\textbf{Single Handover Physical Experiments:} We report success rate, $95\%$ confidence intervals and durations taken over a grid search of the 28 start configurations described in Section~\ref{subsec:assumptions} for the full surgical needle bimanual regrasping task. We report the frequency of three failure modes: (\textbf{P}) error in the presentation procedure, (\textbf{X}) error along the $x$-axis and (\textbf{Y}) error along the $y$-axis. \algname{} significantly outperforms baselines, which either have many presentation failures or grasp positioning failures. The No-Sim-Data ablation also has a high success rate, but we find that the segmentation masks are qualitatively less accurate and have more false positives in the workspace. We present results with \algname{} using three needles (Figure \ref{fig:needle_sizes}) that were unseen in training samples. \algname{} performs best on the larger two needles (2 and 3), and performs worse on the smaller needle 4, where occlusions with the gripper are more severe.}
\label{table:fine_grain_succ}
\begin{center}
\resizebox{\textwidth}{!}{%
\begin{tabular}{|l|r|r|rr|r|rrr|}
\toprule
 & \multicolumn{2}{c|}{\textsc{Success Rate}} & \multicolumn{2}{c|}{\textsc{$95\%$ Conf. Int.}} & \textsc{Completion Time (s)} & \multicolumn{3}{c|}{\textsc{Failures}}\\
 & \multicolumn{1}{c}{Successes / Total} & \multicolumn{1}{c|}{$\%$ Success} & Low & High & & \multicolumn{1}{c}{P} & \multicolumn{1}{c}{X} & \multicolumn{1}{c|}{Y}\\
\midrule
\textsc{Open Loop} & $28/56$ & $50.0$ & $36.3$ & $63.7$ & $\mathbf{14.85 \pm 2.43}$ & $0$ & $0$ & $28$\\
\textsc{Shared $(x, y)$ Grasp Policy} & $2/28$ & $7.1$ & $0.9$ & $23.5$ & $19.41\pm 4.23$ & $0$ & $0$ & $26$\\
\textsc{No Sim Data} & $53/56$ & $94.6$ & $85.1$ &  $98.9$ &$24.58 \pm 3.39$& $3$ & $0$ & $0$ \\
\textsc{Depth-Based Presentation} & $6/28$ & $21.4$ & $8.3$ & $41.0$ & $32.61 \pm 7.42$ & $22$ & $0$ & $0$\\
\textsc{\algname{} (Left to Right)} & $\mathbf{108/112}$ & $\mathbf{96.3}$ & $91.1$ & $99.0$ & $21.45 \pm 3.08$ & $4$ & $0$ & $0$\\
\textsc{\algname{} (Right to Left)} & $\mathbf{108/112}$ & $\mathbf{96.3}$ & $91.1$ &  $99.0$& $23.93 \pm 3.20$ & $1$ & $0$ & $3$\\
\midrule
\multicolumn{9}{|c|}{\textsc{Needles Unseen in Training}}\\
\Xhline{0.5\arrayrulewidth}
\textsc{\algname{} (Right to Left, Needle 2)} & $26/28$ & $92.9$ & $76.5$ & $99.1$ & $23.79 \pm 2.55$ & $1$ & $0$ & $1$\\
\textsc{\algname{} (Right to Left, Needle 3)} & $25/28$ & $89.3$ & $71.8$ & $97.7$ & $23.15 \pm 3.46$ & $3$ & $0$ & $0$\\
\textsc{\algname{} (Right to Left, Needle 4)} & $21/28$ & $75.0$ & $55.1$ & $89.3$ & $23.44 \pm 4.13$ & $5$ & $0$ & $2$\\
\bottomrule
\end{tabular}}
\end{center}
\label{tab:single_handover_results}
\end{table*}
\begin{table*}[t]
\caption {\textbf{Multiple Handover Physical Experiments:} To evaluate the consistency of the \algname{}, we evaluate it on the full multiple handover \algname{} task with maximum handoffs $N_{\rm max} = 50$. We run \algname{} with two different needle orientations during the trial, and report the number of successful handovers (\textbf{Num}), time per handover (\textbf{Time}), and failure mode (\textbf{F}). \algname{} averages 26.20 and 38.60 successful passes in each configuration, and has three runs with no failures.}
\label{table:fine_grain_succ}
\begin{center}
\resizebox{\textwidth}{!}{%
\begin{tabular}{|l|rrr|rrr|rrr|rrr|rrr|rr|}
\toprule
 & \multicolumn{3}{c|}{\textsc{Trial 1}} & \multicolumn{3}{c|}{\textsc{Trial 2}} & \multicolumn{3}{c|}{\textsc{Trial 3}} & \multicolumn{3}{c|}{\textsc{Trial 4}} & \multicolumn{3}{c|}{\textsc{Trial 5}} & \multicolumn{2}{c|}{\textsc{Avg.}}\\
 & \multicolumn{1}{c|}{\textsc{Num}} & \multicolumn{1}{c|}{\textsc{Time}} & \multicolumn{1}{c|}{\textsc{F}}
 & \multicolumn{1}{c|}{\textsc{Num}} & \multicolumn{1}{c|}{\textsc{Time}} & \multicolumn{1}{c|}{\textsc{F}}
 & \multicolumn{1}{c|}{\textsc{Num}} & \multicolumn{1}{c|}{\textsc{Time}} & \multicolumn{1}{c|}{\textsc{F}}
 & \multicolumn{1}{c|}{\textsc{Num}} & \multicolumn{1}{c|}{\textsc{Time}} & \multicolumn{1}{c|}{\textsc{F}}
 & \multicolumn{1}{c|}{\textsc{Num}} & \multicolumn{1}{c|}{\textsc{Time}} & \multicolumn{1}{c|}{\textsc{F}}
 & \multicolumn{1}{c|}{\textsc{Num}} & \multicolumn{1}{c|}{\textsc{T/H}}\\
\bottomrule
Away config. & $12$ & $25.40$ & Y & $45$ & $25.34$ & P & $23$ & $25.42$ & P & $\mathbf{35}$ & $24.71$ & P & $16$ & $25.28$ & X & $26.20$ & $25.13$\\
Towards config. & $\mathbf{50}$ & $\mathbf{26.57}$ & $\mathbf{-}$ & $\mathbf{50}$ & $26.76$ & $-$ & $\mathbf{50}$ & $27.00$ & $-$ & $20$ & $25.08$ & Y & $23$ & $26.48$ & P & $38.60$ & $26.56$\\
\bottomrule
\end{tabular}}
\end{center}
\label{tab:multi_handover_results}
\end{table*}

To evaluate the design choices used in the fine-grained grasping policy, we compare to the following baselines:
\begin{itemize}
    \item \textit{Open Loop:} Executes an open loop grasping motion to grasp the needle based only on needle geometry and inverse kinematics.
    \item \textit{Shared $(x, y)$ Grasp Policy:} Trains a single policy to output $x$ and $y$ displacements, and takes both $I_O(t)$ and $I_L(t)$ as input.
\end{itemize}
To evaluate whether the system can transfer to needles \textit{unseen in training}, we evaluate \algname{} on three additional needles as in Figure~\ref{fig:needle_sizes}.


\subsection{Experimental Setup}
We perform experiments using the daVinci Research Kit (dVRK) surgical robot \cite{dvrk2014}, a cable-driven surgical robot with two needle drivers, a gripper which can open 1cm. 

For perception, the setup includes a Zed Mini stereo camera angled slightly downwards to face the arms, and an overhead Zivid One Plus M camera facing directly down. Stereo images are captured at 2K resolution, and overhead images are captured at 1080p. Locations of the arms relative to each of the cameras is statically calibrated.


\subsubsection{Single handover}
For single handover experiments, we manually vary the orientation of the gripper before each trial to the orientations described in \ref{subsec:assumptions}. A handoff is considered successful if the needle switches from one gripper to the other, and at the end is fully supported by the other gripper.
\subsubsection{Multiple handovers}
For multiple handover experiments, we start the needle in the left gripper in a visible configuration to the camera, so that all errors are a result of handoffs rather than initialization. We evaluate two configurations: one where the needle arc ends facing the stereo camera in the grasping configuration (\textbf{Towards}), and one where it faces the opposite direction (\textbf{Away}). This configuration is typically maintained throughout each multi-handover trial because of the consistency of the needle presentation step.

\subsection{Single Handover Results}
We evaluate \algname{} and baselines on the single handover task in Table~\ref{tab:single_handover_results}, and perform multiple systematic passes over the $28$ starting configurations described in Section~\ref{subsec:assumptions}. We find that \algname{} is able to more reliably perform the task than comparisons, which either experience many presentation errors or many grasp positioning errors.

\subsection{Multiple Handover Results}
We evaluate \algname{} on the multiple handover task with $N_{\rm max} = 50$ with two different starting configurations (Table~\ref{tab:multi_handover_results}). We observe that in the first configuration, the algorithm completes $26.20$ successful handovers on average and $38.60$ in the second. In three trials, no errors occur, and we manually terminate them after $50$ successful handovers.

\subsection{Failure Analysis}\label{subsec:failure_analysis}

\algname{} encounters three failure modes: 
\begin{itemize}
    \item \textbf{P}: Presentation error: the robot fails to present the needle in an orientation that is in the plane of the table with the needle tip pointing toward the grasping arm. This may lead to grasping angles that are unreachable or out of the training distribution for the grasping arm.
    \item \textbf{X}: Grasping positioning error (X): the $x$-axis grasping policy fails to line up with the needle prior to executing the$y$-axis grasping policy.
    \item \textbf{Y}: Grasping positioning error (Y): the  $y$-axis grasping policy fails line up with the needle prior to grasping.
\end{itemize}
We categorize all of the failure modes encountered in Table~\ref{tab:single_handover_results}. We find that the open loop grasping policies are not able to consistently position well for grasping. \algname{} has failures that are evenly distributed across the failure modes.
Grasp policy servoing errors stem mainly from needle configurations that are far outside the distribution seen in training.
Presentation phase failures stem primarily from mis-detection of the needle true tip, either because of incomplete segmentation masks or from drift in robot kinematics causing the most distal needle point to not be the tip. This causes the servoing policy to rotate the needle away from the camera, after which sometimes it loses visibility and fails to bring the needle to the pre-handover pose.
Multi-handoff failures most frequently arise because of subtle imperfections in grasp execution where the needle rotates to a difficult angle. During the subsequent handover the needle can become obstructed by the holding gripper, inhibiting the grasping policy.

\section{Discussion}
In this work we present \algname{}, a problem and an algorithm for reliably completing the bimanual regrasping task on unpainted surgical needles. To our knowledge, this work is the first to study the unmodified variant of the regrasping task.
The main limitations of this approach are its reliance on human demonstrations to learn the grasping policy, and sensitivity to needle and environment appearance. We hypothesize that the former could be mitigated via self-supervised demonstration collection, or by exploring unsupervised methods for fine-tuning behavior cloned policies. Future work will address the latter issue by exploring more powerful network architectures leveraging stereo disparity such as \cite{kollar2021simnet}, and designing more autonomous data collection techniques which can label real needle data without human input. 
In future work, we will also study how to reorient needles between handovers for precise control of needle-in-hand pose
and attempt to make needle tracking more robust to occlusions from tissue phantoms.


\renewcommand*{\bibfont}{\footnotesize}
\printbibliography 

@String { icra    = {IEEE International Conference on Robotics and Automation (ICRA)} }

@String { ieeera  = {IEEE Robotics and Automation Letters (RA-L)} }

@String { iros    = {IEEE/RSJ International Conference on Intelligent Robots and Systems (IROS)} }

@String { ismr    = {International Symposium on Medical Robotics (ISMR)} }

@String { case    = {IEEE Conference on Automation Science and Engineering (CASE)} }

@String { corl    = {Conference on Robot Learning (CoRL)} }

@inproceedings{paradis2020intermittent,
  booktitle={{Intermittent Visual Servoing: Efficiently Learning Policies Robust to Instrument Changes for High-precision Surgical Manipulation}},
  author={Paradis, Samuel and Hwang, Minho and Thananjeyan, Brijen and Ichnowski, Jeffrey and Seita, Daniel and Fer, Danyal and Low, Thomas and Gonzalez, Joseph E and Goldberg, Ken},
  publisher=icra,
  year={2021}
}

@inproceedings{peng2020real,
  booktitle={{Real-time Data Driven Precision Estimator for RAVEN-II Surgical Robot End Effector Position}},
  author={Peng, Haonan and Yang, Xingjian and Su, Yun-Hsuan and Hannaford, Blake},
  publisher=icra,
  year={2020},
}

@inproceedings{saeidi_suturing_icra_2019,
  author = {Saeidi, H and Le, H N D and Opfermann, J D and Leonard, S and Kim, A and Hsieh, M H and Kang, J U and Krieger, A},
  booktitle = {{Autonomous Laparoscopic Robotic Suturing with a Novel Actuated Suturing Tool and 3D Endoscope}},
  publisher = icra,
  Year = {2019}
}

@inproceedings{seita_icra_2018,
  author = {Daniel Seita and Sanjay Krishnan and Roy Fox and Stephen McKinley and John Canny and Kenneth Goldberg},
  booktitle = {{Fast and Reliable Autonomous Surgical Debridement with Cable-Driven Robots Using a Two-Phase Calibration Procedure}},
  publisher = icra,
  Year = {2018}
}

@inproceedings{thananjeyan2017multilateral,
  booktitle={{Multilateral Surgical Pattern Cutting in 2D Orthotropic Gauze with Deep Reinforcement Learning Policies for Tensioning}},
  author={Thananjeyan, Brijen and Garg, Animesh and Krishnan, Sanjay and Chen, Carolyn and Miller, Lauren and Goldberg, Ken},
  publisher=icra,
  year={2017},
}

@inproceedings{sen2016automating,
  Author = {Sen, S. and Garg, A. and Gealy, D. V. and McKinley, S. and Jen, Y. and Goldberg, K.},
  booktitle = {{Automating Multiple-Throw Multilateral Surgical Suturing with a Mechanical Needle Guide and Sequential Convex Optimization}},
  publisher = icra,
  Year = {2016}
}

@inproceedings{murali2015learning,
  booktitle={{Learning by Observation for Surgical Subtasks: Multilateral Cutting of 3D Viscoelastic and 2D Orthotropic Tissue Phantoms}},
  author={Murali, Adithyavairavan and Sen, Siddarth and Kehoe, Ben and Garg, Animesh and McFarland, Seth and Patil, Sachin and Boyd, W Douglas and Lim, Susan and Abbeel, Pieter and Goldberg, Ken},
  publisher = icra,
  year={2015},
}

@inproceedings{dvrk2014,
  booktitle={{An Open-Source Research Kit for the da Vinci Surgical System}},
  author={Kazanzides, P and Chen, Z and Deguet, A and Fischer, G and Taylor, R and DiMaio, S.},
  publisher=icra,
  year=2014
}

@inproceedings{automated_needle_pickup_2018,
  booktitle = {{Automated Pick-up of Suturing Needles for Robotic Surgical Assistance}},
  Author = {D’Ettorre, C and Dwyer, G and Du, X and Chadebecq, F and  Vasconcelos, F and De Momi, E and Stoyanov, D},
  publisher = icra,
  Year = {2018}
}

@inproceedings{thananjeyan2019safety,
  booktitle={{Safety Augmented Value Estimation from Demonstrations (SAVED): Safe Deep Model-Based RL for Sparse Cost Robotic Tasks}},
  author={Thananjeyan, Brijen and Balakrishna, Ashwin and Rosolia, Ugo and Li, Felix and McAllister, Rowan and Gonzalez, Joseph E and Levine, Sergey and Borrelli, Francesco and Goldberg, Ken},
  publisher =ieeera,
  year=2020,
}

@inproceedings{hwang2020efficiently,
  booktitle={{Efficiently Calibrating Cable-Driven Surgical Robots With RGBD Fiducial Sensing and Recurrent Neural Networks}},
  author={Hwang, Minho and Thananjeyan, Brijen and Paradis, Samuel and Seita, Daniel and Ichnowski, Jeffrey and Fer, Danyal and Low, Thomas and Goldberg, Ken},
  publisher=ieeera,
  year={2020}
}

@inproceedings{hwang2020applying,
  author = {Minho Hwang and Daniel Seita and Brijen Thananjeyan and Jeffrey Ichnowski and Samuel Paradis and Danyal Fer and Thomas Low and Ken Goldberg},
  booktitle = {{Applying Depth-Sensing to Automated Surgical Manipulation with a da Vinci Robot}},
  publisher = ismr,
  Year = {2020}
}

@inproceedings{improved_knots_case_2013,
  booktitle = {{Improved Knot-Tying Methods for Autonomous Robot Surgery}},
  author = {Der-Lin Chow and Wyatt Newman},
  publisher = case,
  year = 2013,
}

@inproceedings{extraction_needles_2019,
  booktitle = {{Automated Extraction of Surgical Needles from Tissue Phantoms}},
  author = {Priya Sundaresan and Brijen Thananjeyan and Johnathan Chiu and Danyal Fer and Ken Goldberg},
  publisher = case,
  year = 2019,
}

@inproceedings{mahler2014case,
  Author = {Mahler, J. and Krishnan, S. and Laskey, M. and Sen, S. and Murali, A. and Kehoe, B. and Patil, S. and Wang, J. and Franklin, M. and Abbeel, P. and Goldberg, K.},
  booktitle = {{Learning Accurate Kinematic Control of Cable-Driven Surgical Robots Using Data Cleaning and Gaussian Process Regression.}},
  publisher = case,
  Year = {2014}
}

@inproceedings{QT-Opt,
  booktitle={{QT-Opt: Scalable Deep Reinforcement Learning for Vision-Based Robotic Manipulation}},
  author={Dmitry Kalashnikov and Alex Irpan and Peter Pastor and Julian Ibarz and Alexander Herzog and Eric Jang and Deirdre Quillen and Ethan Holly and Mrinal Kalakrishnan and Vincent Vanhoucke and Sergey Levine},
  publisher=corl,
  year={2018}
}

@inproceedings{auto_peg_transfer_2015,
  booktitle = {{Autonomous Operation in Surgical Robotics}},
  author = {Jacob Rosen and Ji Ma},
  publisher = {Mechanical Engineering},
  volume={137},
  number={9},
  year = 2015,
}

@inproceedings{yip2017robot,
  booktitle={{Robot Autonomy for Surgery}},
  author={Yip, Michael and Das, Nikhil},
  publisher={The Encyclopedia of Medical Robotics},
  year={2017},
}

@article{hutchinson1996tutorial,
  title={A tutorial on visual servo control},
  author={Hutchinson, Seth and Hager, Gregory D and Corke, Peter I},
  journal={IEEE transactions on robotics and automation},
  volume={12},
  number={5},
  pages={651--670},
  year={1996},
  publisher={IEEE}
}

@article{kragic2002survey,
  title={Survey on visual servoing for manipulation},
  author={Kragic, Danica and Christensen, Henrik I and others},
  journal={Computational Vision and Active Perception Laboratory, Fiskartorpsv},
  volume={15},
  pages={2002},
  year={2002},
  publisher={Citeseer}
}

@article{chaumette2006visual,
  title={Visual servo control. I. Basic approaches},
  author={Chaumette, Fran{\c{c}}ois and Hutchinson, Seth},
  journal={IEEE Robotics \& Automation Magazine},
  volume={13},
  number={4},
  pages={82--90},
  year={2006},
  publisher={IEEE}
}

@article{caron2013photometric,
  title={Photometric visual servoing for omnidirectional cameras},
  author={Caron, Guillaume and Marchand, Eric and Mouaddib, El Mustapha},
  journal={Autonomous Robots},
  volume={35},
  number={2},
  pages={177--193},
  year={2013},
  publisher={Springer}
}

@article{levine2018learning,
  title={Learning hand-eye coordination for robotic grasping with deep learning and large-scale data collection},
  author={Levine, Sergey and Pastor, Peter and Krizhevsky, Alex and Ibarz, Julian and Quillen, Deirdre},
  journal={The International Journal of Robotics Research},
  volume={37},
  number={4-5},
  pages={421--436},
  year={2018},
  publisher={SAGE Publications Sage UK: London, England}
}

@inproceedings{ritcher_bloodflow_2020,
  author = {Florian Richter and Shihao Shen and Fei Liu and Jingbin Huang and Emily K. Funk and Ryan K. Orosco and Michael C. Yip},
  booktitle = {{Autonomous Robotic Suction to Clear the Surgical Field for Hemostasis using Image-based Blood Flow Detection}},
  publisher = {arXiv preprint arXiv:2010.08441},
  Year = {2020}
}

@misc{superhuman_dict,
  title = {Definition of superhuman by Merriam Webster},
  howpublished = {\url{https://www.merriam-webster.com/dictionary/superhuman}},
  note = {Accessed: 2021-05-24}
}

@article{chiu2020bimanual,
  title={Bimanual Regrasping for Suture Needles using Reinforcement Learning for Rapid Motion Planning},
  author={Chiu, Zih-Yun and Richter, Florian and Funk, Emily K and Orosco, Ryan K and Yip, Michael C},
  journal={arXiv preprint arXiv:2011.04813},
  year={2020}
}

@article{kollar2021simnet,
  title={SimNet: Enabling Robust Unknown Object Manipulation from Pure Synthetic Data via Stereo},
  author={Kollar, Thomas and Laskey, Michael and Stone, Kevin and Thananjeyan, Brijen and Tjersland, Mark},
  journal={arXiv preprint arXiv:2106.16118},
  year={2021}
}

@article{krishnan2019swirl,
  title={SWIRL: A sequential windowed inverse reinforcement learning algorithm for robot tasks with delayed rewards},
  author={Krishnan, Sanjay and Garg, Animesh and Liaw, Richard and Thananjeyan, Brijen and Miller, Lauren and Pokorny, Florian T and Goldberg, Ken},
  journal={The International Journal of Robotics Research},
  volume={38},
  number={2-3},
  pages={126--145},
  year={2019},
  publisher={SAGE Publications Sage UK: London, England}
}

@inproceedings{kehoe2014autonomous,
  title={Autonomous multilateral debridement with the raven surgical robot},
  author={Kehoe, Ben and Kahn, Gregory and Mahler, Jeffrey and Kim, Jonathan and Lee, Alex and Lee, Anna and Nakagawa, Keisuke and Patil, Sachin and Boyd, W Douglas and Abbeel, Pieter and others},
  booktitle={2014 IEEE International Conference on Robotics and Automation (ICRA)},
  pages={1432--1439},
  year={2014},
  organization={IEEE}
}

@inproceedings{varier2020collaborative,
  title={Collaborative Suturing: A Reinforcement Learning Approach to Automate Hand-off Task in Suturing for Surgical Robots},
  author={Varier, Vignesh Manoj and Rajamani, Dhruv Kool and Goldfarb, Nathaniel and Tavakkolmoghaddam, Farid and Munawar, Adnan and Fischer, Gregory S},
  booktitle={2020 29th IEEE International Conference on Robot and Human Interactive Communication (RO-MAN)},
  pages={1380--1386},
  year={2020},
  organization={IEEE}
}

@article{calli2017yale,
  title={Yale-CMU-Berkeley dataset for robotic manipulation research},
  author={Calli, Berk and Singh, Arjun and Bruce, James and Walsman, Aaron and Konolige, Kurt and Srinivasa, Siddhartha and Abbeel, Pieter and Dollar, Aaron M},
  journal={The International Journal of Robotics Research},
  volume={36},
  number={3},
  pages={261--268},
  year={2017},
  publisher={SAGE Publications Sage UK: London, England}
}

@INPROCEEDINGS{tracik,
  author={Beeson, Patrick and Ames, Barrett},
  booktitle={2015 IEEE-RAS 15th International Conference on Humanoid Robots (Humanoids)}, 
  title={TRAC-IK: An open-source library for improved solving of generic inverse kinematics}, 
  year={2015},
  volume={},
  number={},
  pages={928-935},
  doi={10.1109/HUMANOIDS.2015.7363472}}

@inproceedings{mihaylova2002comparison,
  title={A comparison of decision making criteria and optimization methods for active robotic sensing},
  author={Mihaylova, Lyudmila and Lefebvre, Tine and Bruyninckx, Herman and Gadeyne, Klaas and De Schutter, Joris},
  booktitle={International Conference on Numerical Methods and Applications},
  pages={316--324},
  year={2002},
  organization={Springer}
}

@inproceedings{salaris2017online,
  title={Online optimal active sensing control},
  author={Salaris, Paolo and Spica, Riccardo and Giordano, Paolo Robuffo and Rives, Patrick},
  booktitle={2017 IEEE International Conference on Robotics and Automation (ICRA)},
  pages={672--678},
  year={2017},
  organization={IEEE}
}

@incollection{whitehead1990active,
  title={Active perception and reinforcement learning},
  author={Whitehead, Steven D and Ballard, Dana H},
  booktitle={Machine Learning Proceedings 1990},
  pages={179--188},
  year={1990},
  publisher={Elsevier}
}

@inproceedings{arruda2016active,
  title={Active vision for dexterous grasping of novel objects},
  author={Arruda, Ermano and Wyatt, Jeremy and Kopicki, Marek},
  booktitle={2016 IEEE/RSJ International Conference on Intelligent Robots and Systems (IROS)},
  pages={2881--2888},
  year={2016},
  organization={IEEE}
}

@article{bajcsy1988active,
  title={Active perception},
  author={Bajcsy, Ruzena},
  journal={Proceedings of the IEEE},
  volume={76},
  number={8},
  pages={966--1005},
  year={1988},
  publisher={IEEE}
}


\clearpage

\end{document}